\documentclass[journal]{IEEEtran}

\usepackage{graphics}
\usepackage{graphicx}
\usepackage{algorithm}

\usepackage{subfigure}
\usepackage{amsfonts}
\usepackage{amsmath,amssymb}
\usepackage{xcolor}

\usepackage[noend]{algpseudocode}

\newtheorem{definition}{Definition}
\newtheorem{theorem}{Theorem}

\begin{document}
\title{A Theoretical Analysis of Granulometry-based Roughness Measures on Cartosat DEMs}

\author{K~Nagajothi,~\IEEEmembership{Member,~IEEE,}
        Sravan~Danda,~\IEEEmembership{Member,~IEEE,}
        Aditya~Challa, %~\IEEEmembership{Member,~IEEE,}
        and~B S ~Daya Sagar,~\IEEEmembership{Senior~Member,~IEEE}% <-this % stops a space
\thanks{K. Nagajothi  is with Regional Remote Sensing Centre, Indian Space Research Organisation, Bangalore 560037, India (e-mail: nagajothi\_k@nrsc.gov.in).}% <-this % stops a space
\thanks{Sravan Danda is with Computer Science and Information Systems, APPCAIR, BITS-Pilani Goa Campus, Zuari Nagar, 403726, India (sravan8809@gmail.com)}
\thanks{Aditya Challa is with Computer Science and Automation, Indian Institute of Science, Bangalore, 560012, India (aditya.challa.20@gmail.com)}
\thanks{B. S. D. Sagar is with Systems Science and Informatics Unit, Indian Statistical Institute-Bangalore Centre, 8th Mile, Mysore Road, RV College PO, Bangalore-560059, India (bsdsagar@isibang.ac.in)}}% <-this % stops a space

%% The paper headers

\maketitle
\begin{abstract}
The study of water bodies such as rivers is an important problem in the remote sensing community. A meaningful set of quantitative features reflecting the geophysical properties help us better understand the formation and evolution of rivers. Typically, river sub-basins are analysed using Cartosat Digital Elevation Models (DEMs), obtained at regular time epochs. One of the useful geophysical features of a river sub-basin is that of a roughness measure on DEMs. However, to the best of our knowledge, there is not much literature available on theoretical analysis of roughness measures. In this article, we revisit the roughness measure on DEM data adapted from multiscale granulometries in mathematical morphology, namely \textit{multiscale directional granulometric index (MDGI)}. This measure was classically used to obtain shape-size analysis in greyscale images. In earlier works, MDGIs were introduced to capture the characteristic surficial roughness of a river sub-basin along specific directions. Also, MDGIs can be efficiently computed and are known to be useful features for classification of river sub-basins. In this article, we provide a theoretical analysis of a MDGI. In particular, we characterize non-trivial sufficient conditions on the structure of DEMs under which MDGIs are invariant. These properties are illustrated with some fictitious DEMs. We also provide connections to a discrete derivative of volume of a DEM. Based on these connections, we provide intuition as to why a MDGI is considered a roughness measure. Further, we experimentally illustrate on Lower-Indus, Wardha, and Barmer river sub-basins that the proposed features capture the characteristics of the river sub-basin. %These features may be used in combination with other river basin features for a detailed analysis of the river basins. 
\end{abstract}


% \fi
\IEEEpeerreviewmaketitle

\section{Introduction}

The study of geophysical properties of rivers is an important problem in the remote sensing community. A study of a river sub-basin at regular time epochs over a large span of time helps understand the evolution of the river. The evolution of river sub-basins provides information required to prioritize the rivers that need immediate attention for conservation/ identify natural calamities etc. However, such a study is highly dependent on extracting meaningful geophysical features of the river sub-basins. For example, the complexity of the surficial roughness of a river sub-basin provides information as to what the dominant wind directions are, in that region.   % why is the study of river basin DEMs important

Recall that Cartosat Digital Elevation Models (DEMs) are typically used to compute geophysical features of river sub-basins. In literature, several studies indicate that surficial roughness is an important characteristic of a river sub-basin \cite{turcotte1989fractals,florinsky1998derivation,sagar2007universal,sagar2013mathematical,sagar2004allometric,sagar2004fractal,tay2006allometric,chockalingam2005morphometry,tay2007granulometric}. However, these studies lack a detailed theoretical analysis of the roughness measures proposed. In this article, we analyse in detail, a surficial roughness measure that was proposed in \cite{nagajothi2019classification} i.e. \textit{multiscale directional granulometric index (MDGI)}, a special case of a more general measure namely a \textit{multiscale granulometric index}. % importance of roughness measures and multiscale directional granulometric index in particular

Multiscale granulometric index was originally proposed in \cite{maragos1989pattern} to obtain a shape-size analysis of objects in greyscale images. As greyscale images can be viewed as digital surfaces with the greyscale intensity at each pixel representing the height of the surface, these measures have been adapted to DEMs \cite{tay2007granulometric}. It was shown experimentally that such an adaptation is indeed useful from an application point-of-view i.e. to classify river sub-basins \cite{nagajothi2019classification}. A natural question would then be to ask: \textit{Can we characterize the equivalence classes of DEMs obtained by the equivalence relation - two DEMs are equivalent if their MDGI are identical?} In other words, can we find necessary and sufficient conditions on the structure of a DEM under which a MDGI is invariant? In this article, we partially answer this question and provide theoretical insights on how a MDGI varies with the structure of a DEM. % brief overview of what this article is about (add an image illustration)

 In particular, the contributions of this article are as follows:  
\begin{enumerate}
\item We provide an alternate visualization of the definition of a MDGI proposed in \cite{nagajothi2019classification}.
\item Using the alternate visualization, we characterize non-trivial sufficient conditions on DEMs under which a MDGI is invariant. The invariance properties are intuitively explained and illustrated on fictitious DEMs. 
\item We analyse the relation between a MDGI and a discrete derivative of volume of a DEM. Using this analysis, we provide an intuition as to why a MDGI is considered a roughness measure.
\item Further, a preliminary application of MDGI is shown on real data i.e. Cartosat-1 DEM data of Lower-Indus, Wardha, and Barmer river sub-basins to show that these measures capture characteristics of the sub-basin. 
\end{enumerate} % contributions of the article

The rest of the article is organized as follows: In section \ref{sec:Granulo}, we provide the definitions of basic morphological operators and multiscale granulometric index. Also, the existing literature on the usage of directional granulometric indices is briefly described. Section \ref{sec:Main} contains the core contributions of the article i.e. an alternate visualization of a MDGI, the invariance properties, relation to discrete derivative on the volume of a DEM, and an intuition as to why a MDGI is considered a roughness measure. Section \ref{sec:Experiments} contains experiments on real data i.e. on watersheds of Indus, Wardha and Barmer river sub-basins.  % organization of the rest of the sections

\section{Multiscale Granulometric Index}
\label{sec:Granulo}

In this section, we recall the formal definitions of a multiscale granulometric index and briefly describe the existing literature. First, we start with the basic definitions.

\subsection{Elementary Morphological Operators}
\label{subsec:ElementaryMorph}

%In this subsection, we recall the preliminary notions and definitions that are required to follow the remainder of the article.  

\begin{definition}
Let $A \subset \mathbb{Z}^2$ be a finite set. A digital elevation model (DEM) of a river basin/sub-basin is represented as a function $f: A \rightarrow H$ where $H \subset\mathbb{Z}^{+}$ is a finite set.
\end{definition}

\begin{figure}[h]
	\centering\	
	\includegraphics[width=0.4\columnwidth]{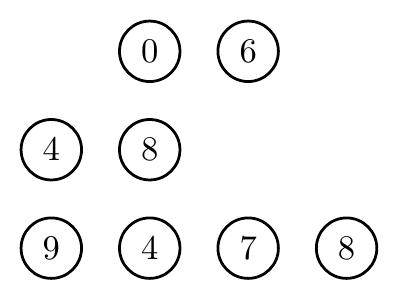} 
	\caption{A non-rectangular fictitious DEM. Each circle represents a physical square area. The values inside each of the circles are the elevations.}
\label{fig:DEM}
\end{figure}

Each point $a \in A$ represents a small physical, square area and $f(a)$ represents the discretized average height of the physical area. Also, observe that $A$ is possibly non-rectangular i.e. $A = \cup_{i=n_1}^{n_2} A_i$ where $A_i = \{(i,j): m_{1,i} \leq j \leq m_{2,i}\}$). See Fig \ref{fig:DEM} for an illustration. This is in contrast to greyscale images where $A$ is always rectangular i.e. $m_{1,i}$ and $m_{2,i}$ are independent of $i$. The number of distinct elements in $H$ are comparable to the number of grey levels in a greyscale image. A higher cardinality of $H$ indicates a finer resolution in the elevations and is analogous to a finer spectral resolution in greyscale images. 

Next, we need the definition of a structuring element. Using the notion of a structuring element, dilation and erosion, the fundamental blocks of roughness measures based on multiscale granulometric index are then defined. Note that we restrict the definition of structuring element i.e. assume that the structuring element contains its origin and is symmetric. This definition suffices for the purposes of this article.

\begin{definition}
A structuring element $SE \subset \mathbb{Z}^{2}$ is a finite set such that: (1) $(0,0) \in SE$, (2) $(i,j) \in SE \Rightarrow (-i,-j) \in SE$.
\end{definition}

\begin{figure}[h]
	\centering\	
	\includegraphics[width=0.8\columnwidth]{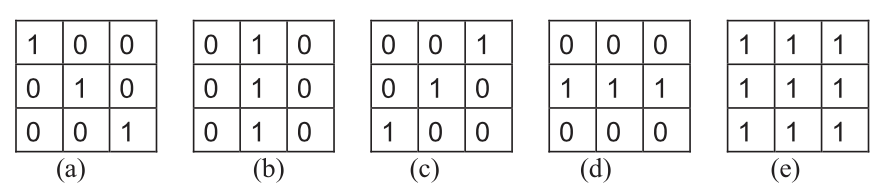} 
	\caption{\textbf{Left to right}: Types of structuring elements used in this article. The centre element refers to $(0,0)$. A value of $1$ indicates that the point corresponding to the coordinates are present in the structuring element and a value of $0$ otherwise.}
\label{fig:SEs}
\end{figure}

The different types of structuring elements used in this article are given by $B_1 = \{(-1,1),(0,0),(1,-1)\}$, $B_2 = \{(0,1),(0,0),(0,-1)\}$, $B_3 = \{(-1,-1),(0,0),(1,1)\}$, $B_4=\{(-1,0),(0,0),(1,0)\}$, and $B = \{(x,y) \in \mathbb{Z}^{2}: -1 \leq x,y \leq 1 \}$. Fig \ref{fig:SEs} provides a pictorial representation of these structuring elements. Observe that each of the structuring elements $B_1,B_2,B_3,B_4$ are $3$ units long and are effectively one-dimensional.

Recall that a greyscale dilation and a greyscale erosion are defined as follows:
\begin{definition}
Let $f: A \rightarrow H$ be a DEM and let $SE$ be a structuring element, then a dilation of $f$ by $SE$ is given by
	\begin{equation}
	\left[f \oplus {SE}\right](x,y) = max_{(s,t) \in {SE}}\{f(x+s,y+t)\}
	\end{equation}
	
where $SE$ is a structuring element	
\end{definition}

\begin{definition}
Let $f: A \rightarrow H$ be a DEM and let $SE$ be a structuring element, then an erosion of $f$ by $SE$ is given by
	\begin{equation}
	\left[f \ominus {SE}\right](x,y) = min_{(s,t) \in {SE}}\{f(x+s,y+t)\}
	\end{equation}
	
where $SE$ is a structuring element	
\end{definition}

Next, we need the definition of a morphological opening and a multiscale morphological opening.

\begin{definition}
Let $f: A \rightarrow H$ be a DEM and let $SE$ be a structuring element, then an opening of $f$ is given by
	\begin{equation}
	\left[f \circ {SE}\right](x,y) = \left[\left[f \ominus {SE} \right] \oplus {SE}\right](x,y)
	\end{equation}

\end{definition} 

\begin{definition}
Let $f: A \rightarrow H$ be a DEM and let $SE$ be a structuring element, then a multiscale opening of $f$ is given by
	\begin{equation}
	\left[f \circ n{SE}\right](x,y) 
	\end{equation}

where $nSE = SE \oplus SE \oplus \cdots \oplus SE$ with the number of dilations in the telescoping expression being $n-1$.
\end{definition} 

% definitions

\subsection{Directional Multiscale Granulometric Index}
\label{subsec:MultiGranulo}

Before we provide a formal definition of a multiscale granulometric index, we need to define the notion of volume of a DEM.

\begin{definition}
Let $f: A \rightarrow H$ be a DEM. The volume of $f$, $V(f)$ is defined as follows:
\begin{equation}
\label{eq:Volume}
V(f) = \sum_{a \in A}f(a)
\end{equation}
\end{definition}

Intuitively, the volume of a DEM captures the physical volume of a DEM on and above the altitude chosen to be zero. For example, the volume of DEM shown in Fig \ref{fig:DEM} is $46$. It is easy to see that an application of a multiscale opening results in a DEM with lower volume as $n$ increases. Also, it is easy to see that there exists $N_0 \in \mathbb{N}$ such that $V(f \circ n{SE}) = V(f \circ (n+1){SE}) \ \forall n \geq N_0$. Recall that the definition of multiscale granulometric index is given by

\begin{definition}
\label{def:GranuloIndex}
\begin{equation}
\label{eq:GranuloIndex1}
GI_{SE}(f) = - \sum_{n=0}^{\infty}p_nlog(p_n)
\end{equation}

where

\begin{equation}
\label{eq:GranuloIndex2}
p_n = \frac{V(f \circ n{SE}) - V(f \circ (n+1){SE})}{V(f)}
\end{equation}
\end{definition}

Note that the existence of $N_0 \in \mathbb{N}$ such that $V(f \circ n{SE}) = V(f \circ (n+1){SE}) \ \forall n \geq N_0$ ensures that the summation is finite. The terms inside the summation for $n \geq N_0$ have to be interpreted as zero. When the structuring element $SE$ is chosen to be one of $B_1,B_2,B_3,B_4$, the obtained multiscale granulometric index is said to be a directional multiscale granulometric index or MDGI. Intuitively, this makes sense as each of $B_1,B_2,B_3,B_4$ are linear and indicate four primary directions.

\subsection{Existing Literature}

Multiscale granulometric index was first introduced in \cite{maragos1989pattern} to perform a shape-size analysis of objects in greyscale images. Then it was used to analyse textures in greyscale images \cite{rzadca1994multivariate}. Later, these ideas were generalized to analyse soil section image analysis \cite{tzafestas2002shape}. Multiscale granulometric index was theoretically analyzed from the perspective of identifying shapes and sizes of objects in greyscale images. The utility of granulometries in greyscale images led to the development of efficient algorithms for specialized classes of structuring elements \cite{vincent2000granulometries,morard2012one}.

Very recently, these ideas were adapted to DEMs. It was experimentally shown in \cite{nagajothi2019classification} that multiscale granulometric indices obtained using specific structuring elements retain characteristic information of the river basins. A natural question would then be to ask: \textit{Can we find necessary and sufficient conditions on the structure of a DEM under which the directional granulometric index is invariant?} In the next section, we partially answer this question by identifying non-trivial sufficient conditions on DEMs such that all DEMs satisfying such conditions have the same directional granulometric index.

\section{Theoretical Analysis of Directional Granulometric Indices}
\label{sec:Main}

In this section, we analyse the MDGIs from a theoretical perspective. Firstly, we recall some modified definitions from graph theory to suit the purposes of subsequent analysis on a MDGI. Secondly, we provide an alternate way to view a MDGI using graphs. Then, by building on this visualization of MDGI, we provide intuition on sufficient conditions under which DEMs have the MDGI. Then, we prove the main result of this article formalizing the intuition i.e. characterization of non-trivial sufficient conditions on structure of a DEM such that the MDGI is invariant. This is followed by a short subsection analysing the relation between MDGI and a discrete derivative of volume of a DEM. Using this analysis, we provide intuition as to why a MDGI is considered a roughness measure.

\subsection{Some Modified Graph Definitions}

\begin{definition}
\label{def:NodeWeightedGraph}
$\mathcal{G} = (V,E,W)$ is said to be a node-weighted graph if $V$ denotes the set of nodes is a finite set, $E \subset \{\{v_i,v_j\}:v_i \neq v_j, \ v_i,v_j \in V\}$ denotes the set of edges, and $W:V \rightarrow H$ is a non-negative integer-valued function on $V$ such that $H \subset \mathbb{Z}^{+}$ is a finite set.
\end{definition}

\begin{figure}[h]
	\centering\	
	\includegraphics[width=0.4\columnwidth]{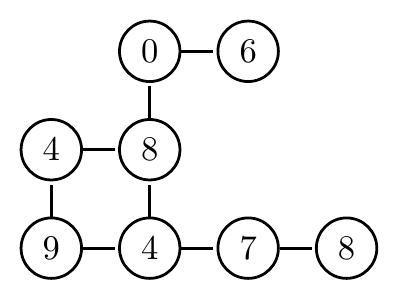} 
	\caption{A node-weighted graph constructed on the DEM given by Fig \ref{fig:DEM}. A $4$-adjacency relation is used to construct the set of edges.}
\label{fig:NWG}
\end{figure}

A node-weighted graph can be used to model DEMs taking into account the spatial relations of neighbouring physical areas on which the elevations are stored. For example, Fig \ref{fig:NWG} shows a node-weighted graph constructed on a fictitious DEM illustrated by Fig \ref{fig:DEM}.

\begin{definition}
Let $\mathcal{G} = (V,E,W)$ be a node-weighted graph. Let $W:V \rightarrow H$ and $h \in H$. $\mathcal{G}_{\geq h} = (V_{\geq h}, E_{\geq h}, W|_{V_{\geq h}})$ is said to be an upper-thresholded subgraph of $\mathcal{G}=(V,E,W)$ at elevation $h$, where $V_{\geq h} = \{v \in V: W(v) \geq h\}$, $E_{\geq h} = \{\{v_i,v_j\}: \{v_i,v_j\} \in E \ and \ W(v_i) \geq h, W(v_j) \geq h\}$, and $W|_{V_{\geq h}}$ denotes the restriction of the function $W$ to $V_{\geq h}$
\end{definition}

\begin{figure}[h]
	\centering\	
	\includegraphics[width=0.4\columnwidth]{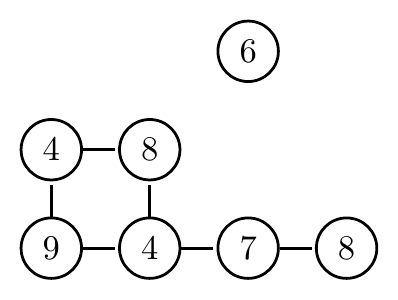} 
	\caption{Upper-thresholded graph constructed on the node-weighted graph given by Fig \ref{fig:NWG}. The threshold is set at elevation $3$.}
\label{fig:UTG}
\end{figure}

Intuitively, an upper-thresholded subgraph of a node-weighted graph constructed on a DEM provides an abstraction of the sub-structure of the the DEM that is above an elevation level. For example, Fig \ref{fig:UTG} illustrates the upper-thresholded subgraph at elevation $3$ on the node-weighted graph given by Fig \ref{fig:NWG}.

\begin{definition}
Let $\mathcal{G} = (V,E,W)$ be a node-weighted graph. A subset of nodes $V_1 \subset V$ is said to be connected if for every pair of nodes $v_s, v_t \in V_1$, there exists a sequence of nodes $<v_s=v_0,v_1,\cdots,v_{r-1},v_r=v_t>$ such that $\{v_i,v_{i+1}\} \in E$ for every $0 \leq i \leq r-1$. A subset of nodes $V_1 \subset V$ is said to be maximally connected if (1) $V_1$ is connected, and (2) $V_1 \subset V_2 \subset V$ and $V_2$ is connected implies $V_2 = V_1$.
\end{definition}

We remark that given any node-weighted graph $\mathcal{G} = (V,E,W)$, the set $V$ can be decomposed uniquely as a disjoint union of maximally connected subsets of $V$. For example, the node-weighted graph in Fig \ref{fig:NWG} has one maximally connected subset which is the vertex set itself. Similarly, the upper-thresholded graph in Fig \ref{fig:UTG} which is also a node-weighted graph has two maximally connected subsets of the vertex set.

\subsection{Another Interpretation of a Directional Granulometric Index}

% horizontal slicing interpretation

Recall from subsection \ref{subsec:MultiGranulo} that a multiscale granulometric index is given by Def \ref{def:GranuloIndex} (see Eq \ref{eq:GranuloIndex1} and Eq \ref{eq:GranuloIndex2}). Intuitively, a multiscale granulometric index measures the entropy of the volume loss on the series of morphological openings with increasing sizes of the structuring element. 

\begin{figure}[h]
	\centering\	
	\includegraphics[width=0.4\columnwidth]{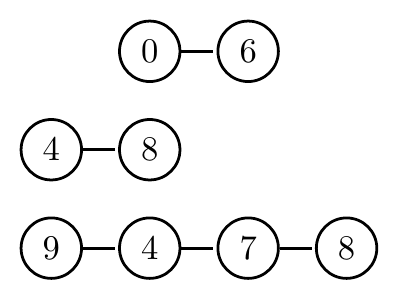} 
	\includegraphics[width=0.4\columnwidth]{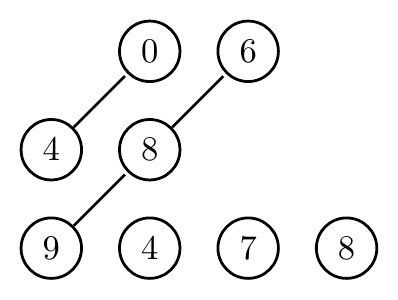} 
	\caption{One dimensional scans on the DEM in Fig \ref{fig:DEM} required to obtain a theoretical analysis of $GI_{B_4}(f)$ and $GI_{B_3}(f)$.}
\label{fig:OneDScan}
\end{figure}

Assume that the structuring element $SE$ is given by one of $B_1, B_2, B_3, B_4$ as defined in subsection \ref{subsec:ElementaryMorph}. Each of the four structuring elements are effectively one-dimensional. Hence, a directional granulometric index effectively measures volume loss on linear scans (but in different directions). Fig \ref{fig:OneDScan} shows an illustration of the scans obtained for SEs $B_4$ and $B_3$. 

Thus, same theoretical analysis on MDGI holds for each of $B_1, B_2,B_3,B_4$. Let $f:A \rightarrow H$ be a DEM. In order to understand the directional granulometric index better, we first try to analyze a one-dimensional DEM i.e. working with DEMs restricting the domain to horizontal scans. The analysis for a generic two-dimensional set $A$ would be a straightforward extension with slightly involved notation. Mathematically, such a restriction would be equivalent to working with sets of type:
 
\begin{equation}
\label{eq:HorizontalSlice}
A_{i_0} = \{(i_0,j): m_1 \leq j \leq m_2 \} \subset A
\end{equation}

for a fixed $i_0 \in \mathbb{Z}$ and $m_1 < m_2 \in \mathbb{Z}$. In general, $m_1,m_2$ depend on $i_0$ as $A$ is not necessarily rectangular. However, we blur this detail to work with a simplified notation. 

We now analyse a MDGI obtained by multiscale openings using horizontal linear structuring elements $\{L_n: n \in \mathbb{Z}^{+}\}$, where $L_n$ denotes a horizontal structuring element with $n$ consecutive $1$s. A similar analysis holds for $\{nB_4: n \in \mathbb{Z}^{+}\}$ because $\{nB_4: n \in \mathbb{Z}^{+}\} \subsetneq \{L_n: n \in \mathbb{Z}^{+}\}$ (in particular $nB_4=L_{2n+1}$ for each $n \in \mathbb{Z}^{+}$). We are now ready to examine $GI_{L_1}(f|_{A_{i_0}})$ given by Eq \ref{eq:GranuloIndex1} and Eq \ref{eq:GranuloIndex2} where $f : A \rightarrow H$ is a DEM. Let $\mathcal{G}^{f} = (A_{i_0},E_{chain},f|_{A_{i_0}})$ denote a node-weighted graph with 

\begin{equation}
\label{eq:ChainGraphEdges}
E_{chain}=\{\{(i_0,j),(i_0,j+1)\}: m_1 \leq j \leq m_2-1\}
\end{equation}
Consider the sequence of upper-thresholded subgraphs of the node-weighted graph $\mathcal{G}^{f} = (A_{i_0},E_{chain},f|_{A_{i_0}})$ at all possible elevations i.e. 

\begin{equation}
\label{eq:UpperThresholdGraphsDEM}
\lbrace\mathcal{G}^{f}_{\geq h}\rbrace_{h=min(H)}^{max(H)} = \lbrace(V^{f}_{\geq h},E^{f}_{\geq h},f|_{V^{f}_{\geq h}})\rbrace_{h=min(H)}^{max(H)} 
\end{equation} 

Let $V^{f}_{\geq h} = \cup_{r=1}^{n_h} V^{f,r}_{\geq h}$ denote the disjoint union of maximally connected subsets for each $h \in [min(H),max(H)]$. Denote $n_{t,h}$ as

\begin{equation}
\label{eq:ConnectedSubFixedLength}
n_{t,h} = |\{V^{f,r}_{\geq h}: |V^{f,r}_{\geq h}|=t\}|
\end{equation}

Here $n_{t,h}$ denotes the number of maximally connected subsets of $V^{f}_{\geq h}$ which are exactly $t$ units long. It is easy to see that the probabilities given by Eq \ref{eq:GranuloIndex2} satisfy

\begin{equation}
\label{eq:GranuloProbsAltView}
p_k \varpropto k\sum_{h=min(H)}^{max(H)}n_{k,h},
\end{equation}

for each $k \in \mathbb{Z}^{+}$. This is because $L_k=kL_1$ is $k$ units long for each $k \in \mathbb{Z}^{+}$. An opening with $kL_1$ removes any maximally connected subset of length less than $k$ units. Hence, probability $p_k$ is proportional to the volume obtained by slices of rectangular blocks that are $k$ units long on the DEM $f$. 

We are now ready to extend these ideas to a generic two-dimensional set $A$. In the two-dimensional case $n_{t,h}$ for each horizontal scan given by Eq \ref{eq:ConnectedSubFixedLength} would be dependent on $i_0$ i.e. the choice of row, denoted by $n_{t,h}^{(i)}$ . Assuming $f:A \rightarrow H$ is the DEM on which we wish to compute the MDGI, Eq \ref{eq:GranuloProbsAltView} would transform to:

\begin{equation}
\label{eq:GranuloProbsAltView2D}
p_k \varpropto \sum_{i=n_1}^{n_2}k\sum_{h=min(H)}^{max(H)}n_{k,h}^{(i)}
\end{equation}   

\subsection{Invariances of Directional Granulometric Indices}

Recall from Sec \ref{sec:Granulo} that a multiscale granulometric index is given by Def \ref{def:GranuloIndex} (see Eq \ref{eq:GranuloIndex1} and Eq \ref{eq:GranuloIndex2}). We are interested to characterize sufficient conditions on the structure of a DEM such that all DEMs that satisfy those conditions have the same MDGI. Mathematically, we need to find non-trivial collections of DEMs $\mathcal{F}_{c} = \{f | GI_{B_4}(f)=c\}$ where $c>0$ is a positive constant. A sufficient condition for a MDGI to be invariant is that the probabilities given by Eq \ref{eq:GranuloIndex2} remain the same. On a closer look at Eq \ref{eq:GranuloProbsAltView}, it is easy to see that if $n_{t,h}$ given by Eq \ref{eq:ConnectedSubFixedLength} remains constant for each $t,h$ then the MDGI for each such DEM is the same. We now state the result formally:

% slide invariance, reflection invariance
\begin{theorem}
\label{th:main}
Let $f_1:A \rightarrow H$ and $f_2:A \rightarrow H$ be distinct DEMs. If the number of maximally connected subsets $n_{t,h}^{(i)}$ (given by Eq \ref{eq:ConnectedSubFixedLength}) of upper-thresholded subgraphs (given by Eq \ref{eq:UpperThresholdGraphsDEM}) for every row $i$, every length $t$, and every elevation $h$ are identical for both $f_1$ and $f_2$ then $ GI_{L_1}(f_1)= GI_{L_1}(f_2)$. 
\end{theorem}

\begin{figure}[h]
	\centering\	
	\includegraphics[width=0.4\columnwidth]{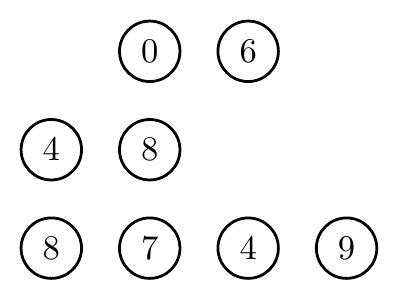} 
	\hspace{1cm}
	\includegraphics[width=0.4\columnwidth]{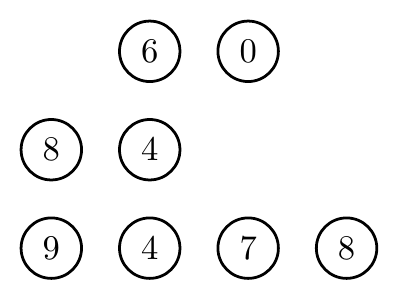} 
	\caption{Two DEMs different from the DEM provided by Fig \ref{fig:DEM} with the same MDGI $GI_{B_4}(.)$.}
\label{fig:InvariantDEMs}
\end{figure}

The proof follows from Eq \ref{eq:GranuloProbsAltView2D}, Eq \ref{eq:GranuloIndex1} and Eq \ref{eq:GranuloIndex2}. To see that the sufficient conditions imposed in Theorem \ref{th:main} are non-trivial, consider $f:A \rightarrow H$. We will now construct a `large' collection of DEMs different from $f$ whose MDGI w.r.t. $B_4$ is identical to that of $f$. Let $A = \cup_{i=n_1}^{n_2} A_i$ where $A_i=\{(i,j): m_{1,i} \leq j \leq m_{2,i} \}$. Define $\hat{f}_{i}: A_i \rightarrow H$ as

\begin{equation}
\label{eq:Reflection}
\hat{f}_{i}(i,j)=f(i,m_{2,i} - j + m_{1,i})
\end{equation}

for each $m_{1,i } \leq j \leq m_{2,i}$. Intuitively, $\hat{f}_{i}$ is the mirror-reflection of $f|_{A_i}$. Now, consider the collection $\mathcal{F}_{Reflection}(f) = \{g:A \rightarrow H: g|_{A_i}=f|_{A_i} \ or \ \hat{f}_{i}\}$. It is easy to see that $|\mathcal{F}_{Reflection}(f)| = 2^{n_2-n_1+1}$. One of the elements of $\mathcal{F}_{Reflection}(f)$ is $f$. Hence, we could construct $2^{n_2-n_1+1} - 1$ different DEMs with the same MDGI. Fig \ref{fig:InvariantDEMs} provides an illustration of this construction i.e. two DEMs different from the DEM provided by Fig \ref{fig:DEM} with same MDGI $GI_{B_4}(.)$.

\begin{figure}[h]
	\centering\	
	\includegraphics[width=0.8\columnwidth]{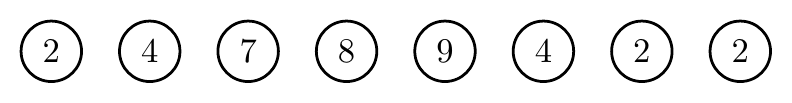} 
	\hspace{1cm}
	\includegraphics[width=0.8\columnwidth]{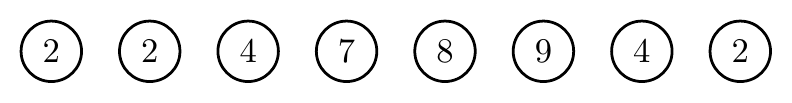} 
	\caption{The top row corresponds to one DEM and the second row corresponds to another DEM. These DEMs have identical $GI_{B_4}(.)$ but they cannot be constructed from each other using the construction provided by Eq \ref{eq:Reflection}.}
\label{fig:InvariantDEMs2}
\end{figure}

In general, it is possible to construct much larger set of DEMs with same MDGI. This is because the domain of DEMs are usually different i.e. arbitrary shaped DEMs are encountered in practice. Also, the conditions characterized by Eq \ref{eq:Reflection} are relatively more restrictive sufficient conditions compared to the sufficient conditions provided by Theorem \ref{th:main}. For example, among the DEMs illustrated in Fig \ref{fig:InvariantDEMs2}, one DEM cannot be constructed from the other using the construction provided by Eq \ref{eq:Reflection}. However, both these DEMs have identical $n_{t,h}$ for each $t,h$ and hence have the same MDGI $GI_{B_4}(.)$. Further, the conditions provided by Theorem \ref{th:main} are sufficient but not necessary.

Next, we discuss another type of invariance which involves transformations on the elevation level i.e. comparing DEMs $f_1:A \rightarrow H_1$ and $f_2: A \rightarrow H_2$ where $H_1$ and $H_2$ are finite subsets of $\mathbb{Z}^{+}$. Formally, we have the following result:

% Depth map resolution: invariance to linear transform
\begin{theorem}
\label{th:ScalingInvariance}
Let $A \subset \mathbb{Z}^{+}$ and $H \subset \mathbb{Z}^{+}$ be finite sets. Assume that $T:H \rightarrow \mathbb{Z}^{+}$ is a scaling transformation i.e. $T(x)=kx$ for some arbitrary but fixed $k \in \mathbb{Z}^{+}$. The DEMs $f : A \rightarrow H$ and $(T \cdot f): A \rightarrow T(H)$ have identical MDGIs i.e. $ GI_{B_l}(f)= GI_{B_l}(T \cdot f)$ for each $1 \leq l \leq 4$. Here $\cdot$ denotes composition of functions.
\end{theorem}

The proof of Theorem \ref{th:ScalingInvariance} follows from the fact that each of the terms on right side of Eq \ref{eq:GranuloProbsAltView2D} is scaled up by the same constant and hence has no effect on the LHS of the same equation.

\subsection{Relation to Discrete Derivative of Volume of a DEM}

In this subsection, we relate a discrete derivative of the volume of a DEM to the MDGI. Firstly, for the sake of simplified notation, assume that we are working on horizontal slices $A_i$ of the domain $A$ of the DEM $f: A \rightarrow H$, i.e. subsets of the type Eq \ref{eq:HorizontalSlice} of $A$. Define

\begin{equation}
\Phi_{h_0}(f|_{A_i}) = \sum_{h = h_0}^{max(H)}\sum_{v \in V^{f,r}_{\geq h} \subset V^{f}_{\geq h}}W(v)
\end{equation}

The quantity $\Phi_{h_0}(f|_{A_i})$ denotes the volume of DEM $f$ on and above elevation $h_0$ on the horizontal slice $A_i$. In particular, $\sum_{i=n_1}^{n_2}\Phi_{min(H)}(f|_{A_i})$ denotes the total volume of DEM $f$, $V(f)$ given by Eq \ref{eq:Volume}. $\Phi_{h_0}(f|_{A_i})$ can be rewritten as
 
\begin{equation}
\Phi_{h_0}(f) = \sum_{h = h_0}^{max(H)} \sum_{t}tn_{t,h}
\end{equation} 

The discrete derivative of the volume of a DEM w.r.t. elevation is hence given by
\begin{equation}
\label{eq:DiscDerivativeOneD}
\Phi_{h_0}(f) - \Phi_{h_0+1}(f) = \sum_{t}tn_{t,h_0}
\end{equation}

Notice that this expression can be interpreted as the sum of areas of maximally connected subsets of $V^{f}_{\geq h_0}$ on the slice $A_i$. In general, when we consider a two-dimensional domain $A$ of the DEM $f$, Eq \ref{eq:DiscDerivativeOneD} transforms to:

\begin{equation}
\label{eq:DiscDerivativeTwoD}
\Phi_{h_0}(f) - \Phi_{h_0+1}(f) = \sum_{i=n_1}^{n_2}\sum_{t}tn_{t,h_0}^{(i)}
\end{equation}
 
\subsection{Why is a Multiscale Directional Granulometric Index considered a Roughness Measure?}

% single peak characterization of constant granulometric index
In this subsection, we provide an intuition as to why a MDGI is regarded as a roughness measure. To accomplish this, we consider a special class of DEMs given by:

\begin{equation}
\mathcal{F}_{UniPeak}= \{f: A \rightarrow H: \sum_{t}n_{t,h} = 1 \forall  \ h \in f(A) \},
\end{equation}

where $A$ denotes a one-dimensional set i.e. $\exists m_1 \leq m_2 \in \mathbb{Z}$ such that $A = \{(i_0,j) | m_1 \leq j \leq m_2\}$. It is easy to see that in such a case the set of discrete derivatives given by Eq \ref{eq:DiscDerivativeOneD} would effectively be a permutation of the probabilities given by Eq \ref{eq:GranuloProbsAltView}. This means that the entropy calculated on the successive discrete derivatives of volume of a DEM is identical to the MDGI when computed on a one-dimensional uni-peak DEM.

\section{Experiments}
\label{sec:Experiments}

In this section, we provide empirical evidence to show that MDGIs capture the characteristic features of a river sub-basin. 

\subsection{Study Area and Data Used}

\begin{figure}[h]
	\centering\	
	\includegraphics[width=1.0\columnwidth]{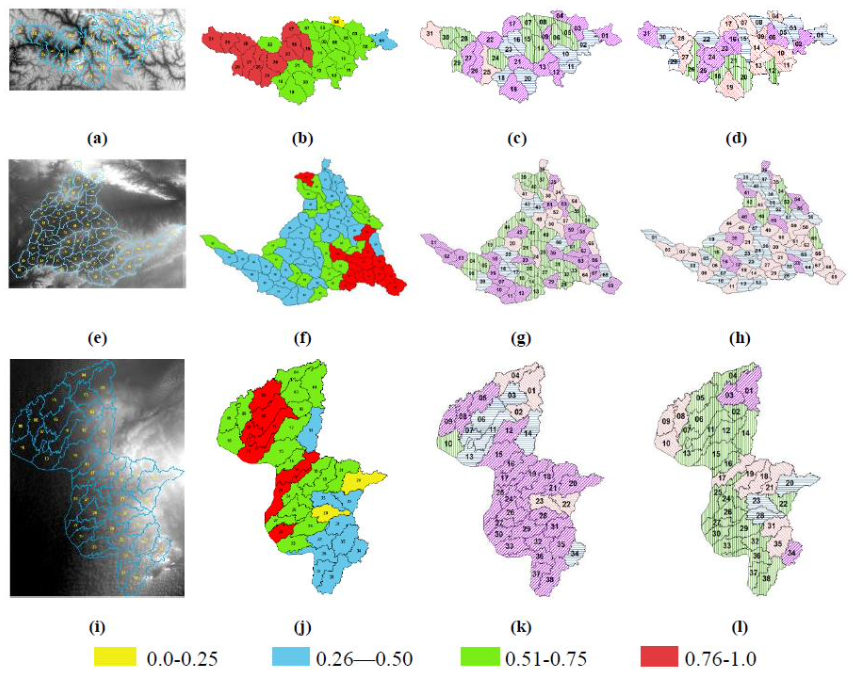} 
	\caption{\textbf{(a), (e), and (i)} Cartosat DEMs of Lower Indus, Wardha, and Barmer sub-basins. The delineations highlight distinct watersheds within each of the sub-basins. \textbf{(b), (f), and (j)} Watersheds classified based on normalized multiscale directional granulometric indices. The MDGIs are all scaled down within each sub-basin and are color-coded as per the ranges mentioned in the legend. The ranges are arbitrarily chosen. \textbf{(c), (g), and (k)} high directional granulometric indices with the colors and the texture highlighting the corresponding SE for which the MDGI is highest, and \textbf{(d), (h), and (l)} low directional granulometric indices with the colors and the texture highlighting the corresponding SE for which the MDGI is lowest, of $31$ watersheds of the Lower Indus, $69$ watersheds of Wardha and $38$ watersheds of Barmer sub-basins respectively. }
\label{fig:RiverBasins}
\end{figure}

We consider lower Indus sub-basin (fluvial), Wardha sub-basin (floodplain), and Barmer sub-basin (desert) for the experiments. Lower Indus sub-basin, one of the $14$ sub-basins lies in between the geographical coordinates of $73^{\circ} 11’$ to $76^{\circ} 44’$ East longitudes and $34^{\circ}42'$ to $36^{\circ} 9'$ North latitudes, is divided into $31$ watersheds of sizes ranging between $319$ sq.km and $1270$ sq.km. Wardha sub-basin, one of the principal tributaries of Godavari river, is situated in between the geographical coordinates of $19^{\circ}18’N$ and $21^{\circ}58’N$ latitudes, and $77^{\circ}20’E$ and $79^{\circ}45’E$ longitudes. This sub-basin has $69$ watersheds. Barmer is another sub-basin of Indus Basin situated between $69^{\circ}48’$ and $71^{\circ}43’$ East longitudes, and between $25^{\circ}28’$ to $27^{\circ}69’$ North latitudes. It is fully under Thar Desert and is divided into $38$ watersheds. Cartosat DEMs of the Lower Indus sub-basins, Wardha sub-basins, and Barmer sub-basins are illustrated in Fig \ref{fig:RiverBasins}.

\subsection{Some Preliminary Observations}

\begin{figure}[h]
	\centering\	
	\includegraphics[width=1.0\columnwidth]{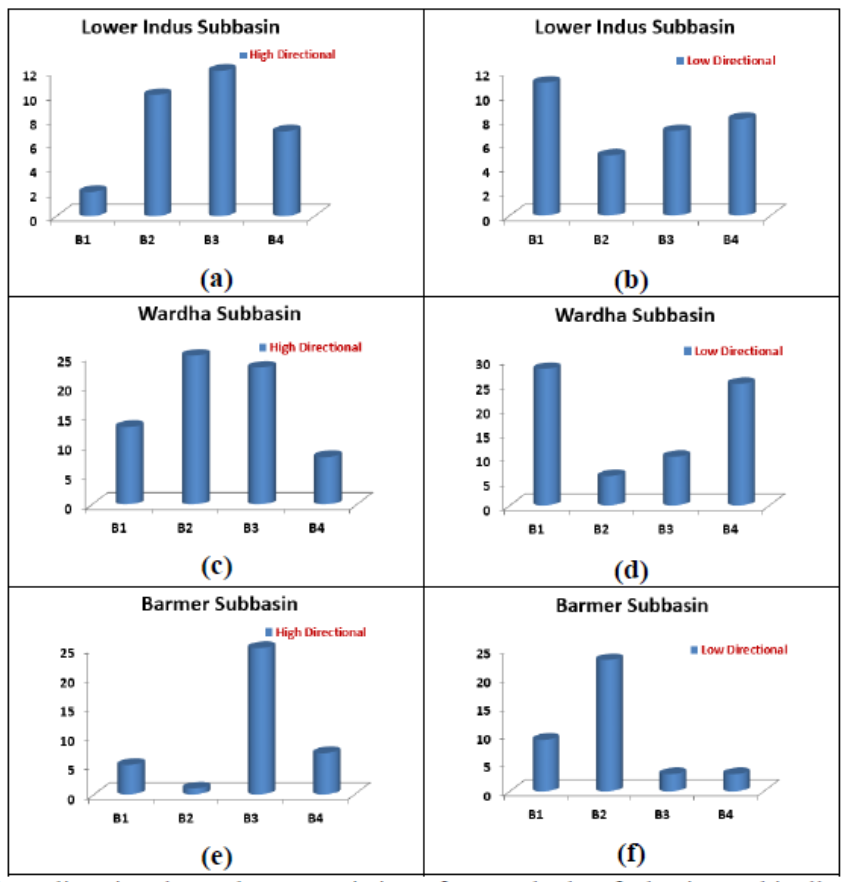} 
	\caption{\textbf{(a), (c), and (e)}: The histograms of the highest order-statistic among normalized HDGIs obtained with $B_1, B_2, B_3, B_4$ as the structuring elements on Lower-Indus, Wardha, and Barmer sub-basins respectively. \textbf{(b), (d), and (f)}: The histograms of the lowest order-statistic among normalized HDGIs obtained with $B_1,B_2,B_3,B_4$ as the structuring elements on Lower-Indus, Wardha, and Barmer sub-basins respectively.}
\label{fig:OrderStatistics}
\end{figure}

We compute the MDGIs of all watersheds in each of the river sub-basins using the structuring elements $B_1,B_2,B_3,B_4$. As there is a variability in the size of each of the watersheds i.e. the domain of each DEM is of different cardinalities, we normalize these MDGIs with the multiscale granulometric index obtained by using $B$ as the structuring element. As a first observation, for each river, we check the order-statistics of the normalized MDGIs. Fig \ref{fig:OrderStatistics} shows a plot of the histograms of the maximum and minimum of the normalized MDGIs among the four primary directions across the river sub-basins. These histograms can be viewed as empirical probability distributions. The maximum (respectively minimum) among the normalized MDGIs is denoted as high directional (respectively low directional) in Figs \ref{fig:RiverBasins} and \ref{fig:OrderStatistics}. It is easy to see that the Barmer sub-basins show a different pattern in the order-statistics as illustrated in Fig \ref{fig:OrderStatistics}. In particular, it is often the case that the maximum is along the direction of $B_3$ and the lowest is given by $B_2$ which is not the case with the other two river sub-basins. However, these order-statistics do not help in identifying the differences between Wardha and Lower-Indus sub-basins.

\subsection{More Observations}

\begin{figure}[h]
	\centering\	
	\includegraphics[width=1.0\columnwidth]{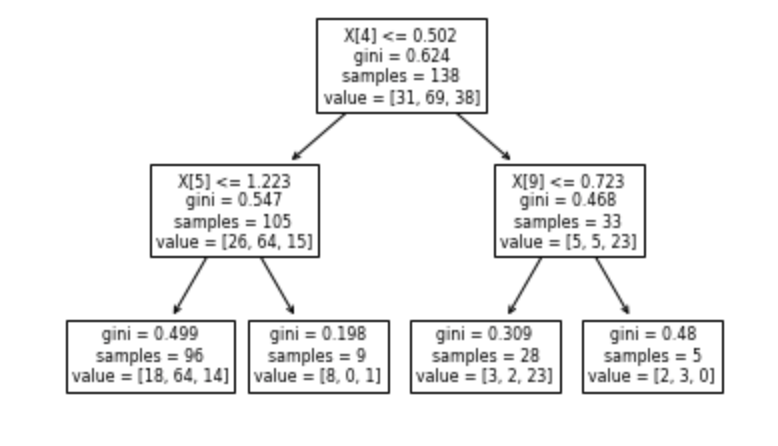} 
	\caption{A decision tree with a depth of $2$ is built using the features $X[0], \cdots, X[15]$ based on the MDGIs as described in the text. This decision tree is the smallest possible classification tree that is capable of separating three classes. This classification tree obtains a classification accuracy of $\approx 71\%$ while a random classifier can obtain a maximum classification accuracy of $50\%$ on classifying $138$ watersheds.}
\label{fig:DecisionTree}
\end{figure}

To identify the differences between the watersheds of all three river sub-basins, we construct features $X[0], \cdots, X[15]$ based on the normalized MDGIs. The details of the construction are as follows:

Let $GI_{B_i}(f)/GI_{B}(f) = Z_i$ for $i=1,2,3,4$
\begin{equation}
X[{(i-1)4+(j-1)}] = 
\begin{cases}
& Z_i \ if \  Z_{(j)}=Z_i \\
& 0 \ otherwise
\end{cases}
\end{equation}

for each $i=1,2,3,4$ and $j=1,2,3,4$, where $Z_{(j)}$ denotes the $j^{th}$ order-statistic among $Z_1,Z_2,Z_3,Z_4$ i.e. $Z_{(1)} \leq Z_{(2)} \leq Z_{(3)} \leq Z_{(4)}$ form a permutation of $Z_1,Z_2,Z_3,Z_4$. As the number of watersheds ($138$ altogether) is small, we do not split the data into training and test sets. Instead, we try to obtain interpretable rules that can classify the watersheds into appropriate sub-basins based on the constructed features. A simple decision tree with a depth of $2$ is constructed. The depth of the decision tree is restricted so as to ensure that we do not over fit the data. Also, the reason for choosing $2$ as the depth is that the minimum depth of a decision tree required to classify $3$ classes is $2$. We observe that such a decision tree (see Fig \ref{fig:DecisionTree}) is capable of obtaining $\approx 71\%$ accuracy. Note that a random classifier on the other hand can obtain a maximum accuracy of $69/138 = 50\%$. This is because the data is class-unbalanced with $31$ Indus, $69$ Wardha and $38$ Barmer watersheds. 

%Nagajothi data  - Decision tree interpretation

This experiment indicates that the features computed from MDGIs carry characteristic information of the sub-basins. To explore the discernibility of MDGI-based features, we built decision trees of larger depths until a maximum depth of $9$. We observed that the accuracies of decision trees of depths $5$, $6$, and $9$ are $\approx 86\%$, $\approx 89\%$, and $\approx 94\%$ respectively. This indicates that MDGI-based features are useful for river sub-basin classification. However, these features cannot be used as standalone features to classify sub-basins. In order to build better classifiers, one needs to incorporate domain knowledge on the river sub-basins through some form of remotely-sensed data or otherwise.

\section{Conclusions and Perspectives}
In this article, we revisit the roughness measure on DEM data adapted from multiscale granulometries in mathematical morphology, namely multiscale directional granulometric index (MDGI). In earlier works, MDGIs were introduced to capture the characteristic surficial roughness of a river sub-basin along specific directions. They are known to be useful features for classification of river sub-basins. In this article, we provided a theoretical analysis of a MDGI. In particular, we characterized non-trivial sufficient conditions on the structure of DEMs under which MDGIs are invariant. These properties are illustrated with some fictitious DEMs. We also provided connections to a discrete derivative of volume of a DEM. Based on these connections, we provided intuition as to why a MDGI is considered a roughness measure. Further, we experimentally illustrated on Lower-Indus, Wardha, and Barmer river sub-basins that the proposed features capture the characteristics of the river sub-basin.

Building on the ideas from this article, one can explore at least two directions: 1) building on the main theorem, one can investigate more sufficient conditions ultimately trying to characterize sufficient and necessary conditions on the structure of a DEM such that MDGI is invariant, 2) on the experimental side, use the features proposed in the article alongside other features on river sub-basins to build better classifiers.

%\section{Temporary}
%
%Figures: Definition 1, 7, 10, Eqn 8, 14, before Theorem 2 outside the characterized conditions one peak example, section 3 E discrete derivative single peak intuition

%\appendices
%% you can choose not to have a title for an appendix
%% if you want by leaving the argument blank
%\section{}

\section*{Acknowledgment}

Nagajothi Kannan would like to thank Regional Remote Sensing Centre, Indian Space Research Organisation. Sravan Danda would like to acknowledge the funding received from BPGC/RIG/2020-21/11-2020/01 (Research Initiation Grant provided by BITS-Pilani K K Birla Goa Campus) and thank APPCAIR, and Computer Science and Information Systems, BITS-Pilani Goa.  Aditya Challa would like to thank Indian Institute of Science (IISc) for the Raman Post Doctoral fellowship. The work of B. S. D. Sagar was supported by the DST-ITPAR-Phase-IV project under the Grant number INT/Italy/ITPAR-IV/Telecommunication/2018.

\bibliographystyle{plain}

\bibliography{references}

\begin{IEEEbiography}[{\includegraphics[width=1in,height=1.25in,clip,keepaspectratio]{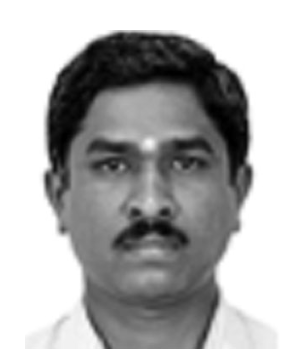}}]{Nagajothi Kannan}
(M’18) received the B.Sc. degree in physics from Bharathiar University, Coimbatore, India, in 1986, and the master’s degree in computer applications from Indira Gandhi National Open University, New Delhi, India, in 2002. He is currently a Senior Scientist with the Indian Space Research Organisation, Bangalore, India. He has more than 20 years of work experience in the area of digital image processing, geographical information system, software development, and system management. He has worked on geospatial application projects of national missions and large-scale user projects that involved design and development of large database, system integration, and customized solutions. His research interests include digital image processing, mathematical morphology, GISci and geospatial applications, and decision support systems. Mr. Nagajothi is a member of IEEE GRSS.
\end{IEEEbiography}

\begin{IEEEbiography}[{\includegraphics[width=1in,height=1.25in,clip,keepaspectratio]{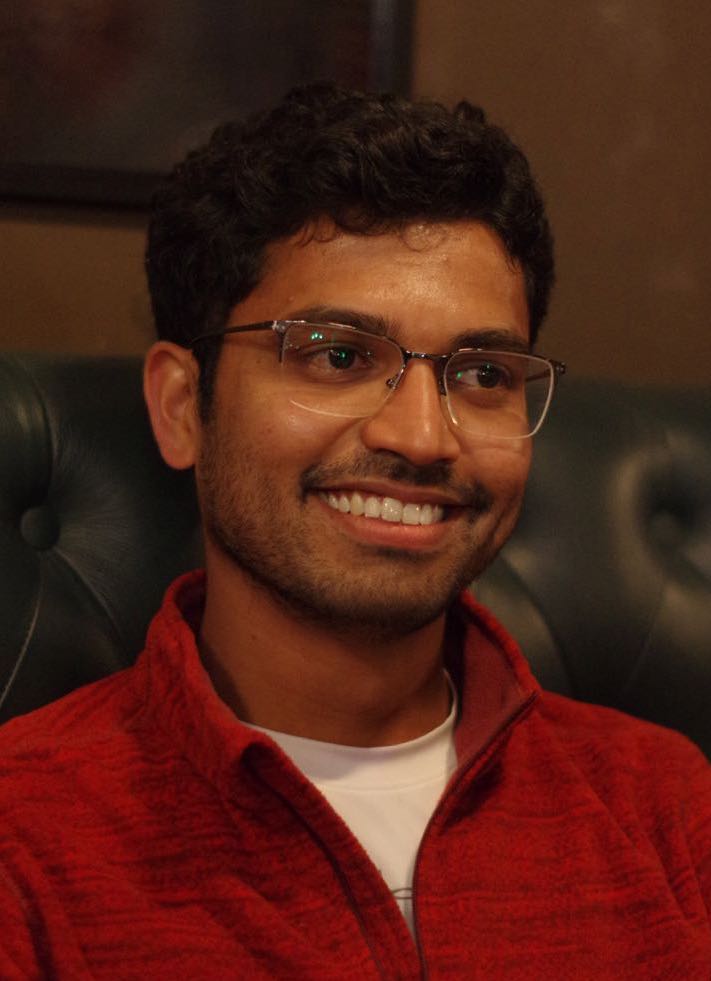}}]{Sravan Danda}
 received the B.Math.(Hons.) degree in Mathematics from the Indian Statistical Institute - Bangalore, and the M.Stat. degree in Mathematical Statistics from the Indian Statistical Institute - Kolkata, in 2009, and 2011, respectively. From 2011 to 2013, he worked as a Business Analyst at Genpact - Retail Analytics, Bangalore. He completed his PhD in computer science from Systems Science and Informatics Unit, Indian Statistical Institute - Bangalore in 2019 under the joint supervision of B.S.Daya Sagar and Laurent Najman. He is currently working as an Assistant Professor at Department of Computer Science and Information Systems, BITS Pilani K K Birla Goa Campus. His current research interests are Discrete Mathematical Morphology and Discrete Optimization in Machine Learning.
\end{IEEEbiography}

\begin{IEEEbiography}[{\includegraphics[width=1in,height=1.25in,clip,keepaspectratio]{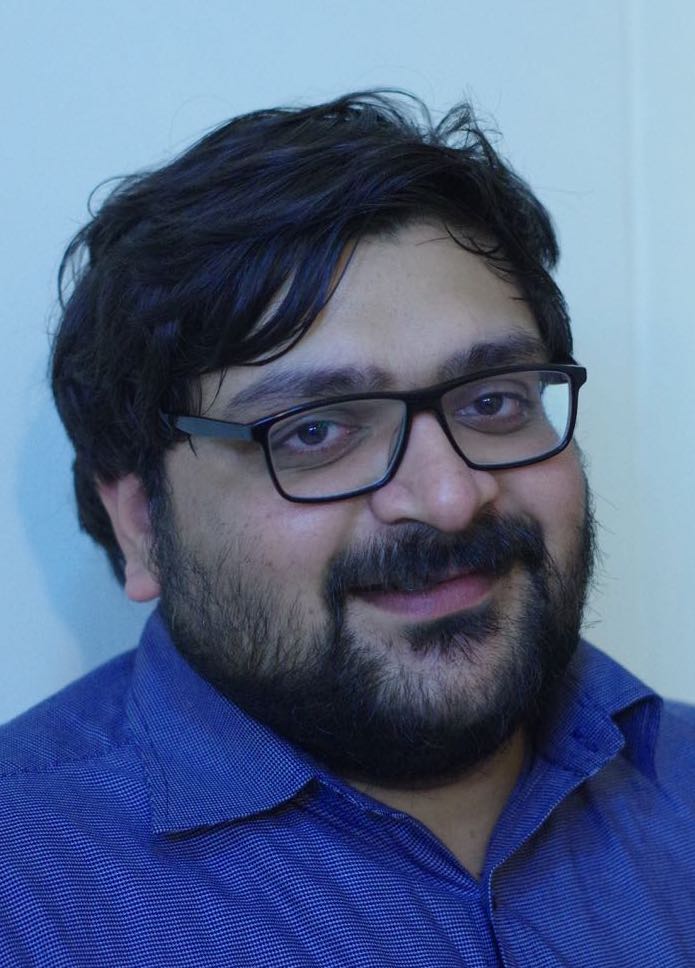}}]{Aditya Challa}
received the B.Math.(Hons.) degree in Mathematics from the Indian Statistical Institute - Bangalore, and Masters in Complex Systems from University of Warwick, UK - in 2010, and 2012, respectively. From 2012 to 2014, he worked as a Business Analyst at Tata Consultancy Services, Bangalore. He completed his PhD in computer science from Systems Science and Informatics Unit, Indian Statistical Institute - Bangalore in 2019. He is currently a Raman PostDoc Fellow at Indian Institute of Science, Bangalore. His current research interests focus on using techniques from Mathematical Morphology in Machine Learning.
\end{IEEEbiography}

\begin{IEEEbiography}[{\includegraphics[width=1in,height=1.25in,clip,keepaspectratio]{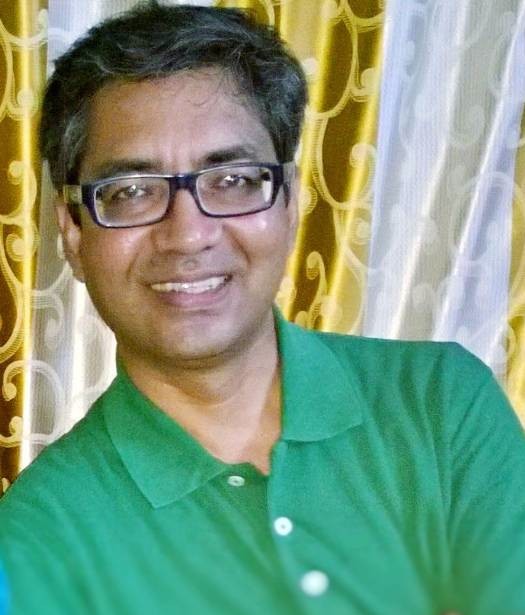}}]{B S Daya Sagar}
 (M'03-SM'03) is a Full Professor of the Systems Science and Informatics Unit (SSIU) at the Indian Statistical Institute. Sagar received his MSc and Ph.D. degrees in Geoengineering and Remote Sensing from the Faculty of Engineering, Andhra University, Visakhapatnam, India, in 1991 and 1994 respectively. He is also the first Head of the SSIU. Earlier, he worked in the College of Engineering, Andhra University, and Centre for Remote Imaging Sensing and Processing (CRISP), The National University of Singapore in various positions during 1992-2001. He served as Associate Professor and Researcher in the Faculty of Engineering \& Technology (FET), Multimedia University, Malaysia, during 2001-2007. Sagar has made significant contributions to the field of geosciences, with special emphasis on the development of spatial algorithms meant for geo-pattern retrieval, analysis, reasoning, modeling, and visualization by using concepts of mathematical morphology and fractal geometry. He has published over 85 papers in journals and has authored and/or guest-edited 11 books and/or special theme issues for journals. He recently authored a book entitled `Mathematical Morphology in Geomorphology and GISci,' CRC Press: Boca Raton, 2013, p. 546. He recently co-edited two special issues on ”Filtering and Segmentation with Mathematical Morphology” for IEEE Journal of Selected Topics in Signal Processing (v. 6, no. 7, p. 737-886, 2012), and ”Applied Earth Observation and Remote Sensing in India” for IEEE Journal of Selected Topics in Applied Earth Observation and Remote Sensing (v. 10, no. 12, p. 5149-5328, 2017). His recent book “Handbook of Mathematical Geosciences”, Springer Publishers, p. 942, 2018 reached 750000 downloads. He was elected as a member of the New York Academy of Sciences in 1995, as a Fellow of the Royal Geographical Society in 2000, as a Senior Member of the IEEE Geoscience and Remote Sensing Society in 2003, as a Fellow of the Indian Geophysical Union in 2011. He is also a member of the American Geophysical Union since 2004, and a life member of the International Association for Mathematical Geosciences (IAMG). He delivered the ”Curzon \& Co - Seshachalam Lecture - 2009” at Sarada Ranganathan Endowment Lectures (SRELS), Bangalore, and the ”Frank Harary Endowment Lecture - 2019” at International Conference on Discrete Mathematics - 2019 (ICDM - 2019). He was awarded the ’Dr. Balakrishna Memorial Award’ of the Andhra Pradesh Academy of Sciences in 1995, the Krishnan Medal of the Indian Geophysical Union in 2002, the ’Georges Matheron Award - 2011 with Lectureship’ of the IAMG, and the Award of IAMG Certificate of Appreciation - 2018. He is the Founding Chairman of the Bangalore Section IEEE GRSS Chapter. He has been recently appointed as an IEEE Geoscience and Remote Sensing Society (GRSS) Distinguished Lecturer (DL) for a two-year period from 2020 to 2022. He is on the Editorial Boards of Computers \& Geosciences, Frontiers: Environmental Informatics, and Mathematical Geosciences. He is also the Editor-In-Chief of the Springer Publishers’ Encyclopedia of Mathematical Geosciences.
\end{IEEEbiography}


%\vfill


\end{document}